%% file: sample-sigconf.tex
\documentclass[sigconf]{acmart}

\usepackage{booktabs} 
\usepackage{graphicx}
\usepackage{subfigure}
\usepackage{makecell}
\usepackage{bbding}
\usepackage{enumitem}
\usepackage{setspace}

\setcopyright{rightsretained}

\copyrightyear{2018} 
\acmYear{2018} 
\setcopyright{acmcopyright}
\acmConference[MM '18]{2018 ACM Multimedia Conference}{October 22--26, 2018}{Seoul, Republic of Korea}
\acmBooktitle{2018 ACM Multimedia Conference (MM '18), October 22--26, 2018, Seoul, Republic of Korea}
\acmPrice{15.00}
\acmDOI{10.1145/3240508.3240550}
\acmISBN{978-1-4503-5665-7/18/10}

\begin{document}

\title{Attribute-Aware Attention Model for Fine-grained Representation Learning}

\author{Kai Han$^{1,*}$, ~Jianyuan Guo$^{1,*}$, ~Chao Zhang$^{1,\star}$, ~Mingjian Zhu$^{1}$}
\affiliation{
  \institution{$^*$Equal contribution, ${^ \star}$Corresponding author\\
  $^{1}$Key Laboratory of Machine Perception (MOE), School of EECS, Peking University, Beijing, China}
}
\email{{hankai, jyguo, c.zhang, zhumingjian}@pku.edu.cn}

\begin{abstract}
How to learn a discriminative fine-grained representation is a key point in many computer vision applications, such as person re-identification, fine-grained classification, fine-grained image retrieval, etc. Most of the previous methods focus on learning metrics or ensemble to derive better global representation, which are usually lack of local information. Based on the considerations above, we propose a novel Attribute-Aware Attention Model ($A^3M$), which can learn local attribute representation and global category representation simultaneously in an end-to-end manner. The proposed model contains two attention models: attribute-guided attention module uses attribute information to help select category features in different regions, at the same time, category-guided attention module selects local features of different attributes with the help of category cues. Through this attribute-category reciprocal process, local and global features benefit from each other. Finally, the resulting feature contains more intrinsic information for image recognition instead of the noisy and irrelevant features. Extensive experiments conducted on Market-1501, CompCars, CUB-200-2011 and CARS196 demonstrate the effectiveness of our $A^3M$. Code is available at \url{https://github.com/iamhankai/attribute-aware-attention}.
\end{abstract}



\maketitle

\begin{spacing}{0.85 }
{\small
\textbf{ACM Reference Format:}\\
Kai Han, Jianyuan Guo, Chao Zhang, Mingjian Zhu. 2018. Attribute-
Aware Attention Model for Fine-grained Representation Learning. In \emph{2018
ACM Multimedia Conference (MM '18), October 22-26, 2018, Seoul, Republic of
Korea.} ACM, New York, NY, USA, 9 pages. https://doi.org/10.1145/3240508.
3240550
}
\end{spacing}
\vspace{0cm}

\input{samplebody-conf}

\nocite{han2018autoencoder}
\bibliographystyle{ACM-Reference-Format}
\bibliography{sample-bibliography}

\bibliographystyle{ACM-Reference-Format}

\end{document}

%% file: samplebody-conf.tex
\section{Introduction}

\begin{figure}[!htp]
\begin{center}
   \includegraphics[width=1.2\linewidth]{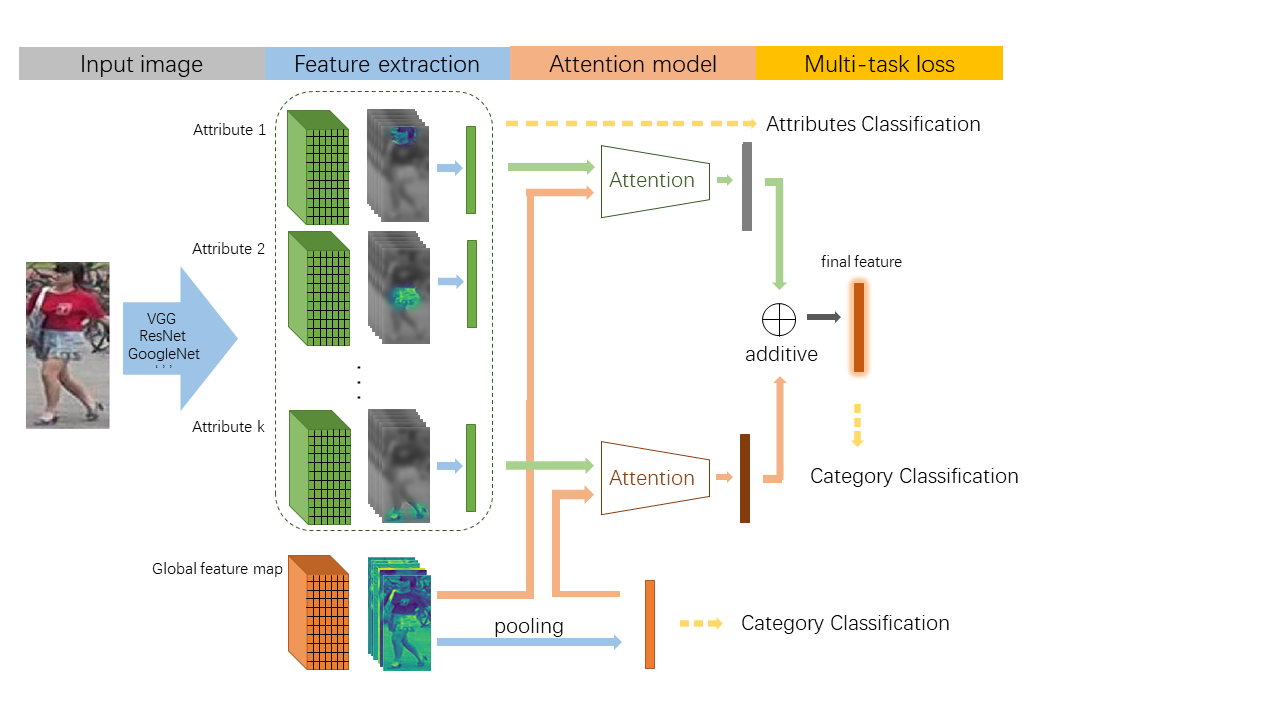}
\end{center}
   \caption{The overall architecture of our Attribute-Aware Attention Model ($A^3M$). Attributes like hair, handbag, shoes, etc. provide rich information for fine-grained feature learning. $A^3M$ learns local attribute feature (green part) and global category feature (orange part) simultaneously in an end-to-end manner. Details of the attention models will be presented in section 3.}
\label{fig:fig1}
\end{figure}

\begin{figure}[!htp]
\begin{center}
   \includegraphics[width=0.75\linewidth]{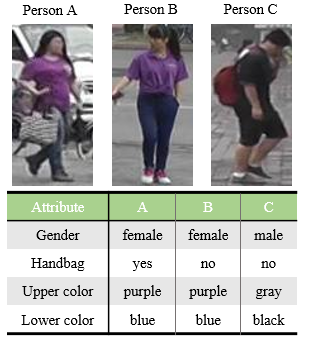}
\end{center}
   \caption{Person image examples with some attributes.}
\label{fig:person}
\end{figure}

Fine-grained visual recognition tasks such as classifying subordinate categories of birds, or identifying a specified person in the surveillance system (see examples in Figure \ref{fig:person}), are quite challenging since the visual differences among categories or instances are so small that they are easily overwhelmed by other factors, such as, object location, pose, lighting or viewpoint changes. In the wave of deep learning, fine-grained visual visual recognition is more and more close to practical application in recent years. A common approach to mining the visual cues in fine-grained scenarios is localizing the manually defined parts and modeling based on these parts \cite{huang2016part}. Recent studies show that this method has improved significantly over simple convolutional neural networks (CNNs). However, most of these works are trained by fully annotated parts, which is time-consuming and laborious. Moreover, the hand-annotated parts may be suboptimal for image recognition. Another approach is to localize the hidden semantic parts only under the supervision of object labels \cite{oquab2015object, durand2016weldon}. Nevertheless, this manner is not reliable enough to mark accurate parts due to weak labels and lack of other helpful information.

Attributes which provide rich information to learn the correlation among categories are important auxiliary signals in image recognition \cite{lampert2009learning,wang2009joint}. Attributes usually describe the high-level properties, which are discriminative for the objects. Taking images in Figure \ref{fig:person} as an example, there are 3 persons, namely, A, B, C taken from Market-1501 dataset \cite{Zheng2015Scalable}. A is similiar to B on the whole, but A carries a bag in hand while B does not, so the attribute handbag is one of the keys to distinguish them. Certainly, other attributes are useful too, such as gender or upper color can help recognizing A from C. Hence, it is crucial to make good use of these local details for fine-grained recognition.

In this paper, we propose an Attribute-Aware Attention Model ($A^3M$) to learn a discriminative representation for fine-grained recognition, object re-identification, etc. As shown in Figure \ref{fig:fig1}, our model includes two branches: attribute feature learning and category feature learning. Our intuition is that small attribute cues are usually crucial to distinguish different categories, such as birds from two species with small difference. Through attention models, category features are used for learning basic representation and the attribute features help to refine category features simultaneously. In fact, the importance of different attributes should vary in different categories, these two attention models can also learn to choose the important attribute features for corresponding input images. The addition of the category features and attribute features helps to gain better representation, and yields better performance.

Main contributions of this work are as follows:
\begin{itemize}
\item We propose Attribute-Aware Attention Model ($A^3M$) to train an attribute-category reciprocal CNN, where attention models can use attribute information to help select key features for person re-identification and fine-grained recognition. Moreover, we use $A^3M$ in fine-grained image retrieval tasks, in which no attribute labels but category information is provided, to combine local and global feature, leading to a better representation without time-consuming ensemble or complicated losses.
\item Empirical results show the superiority of the proposed $A^3M$ for person re-identification, fine-grained classification and fine-grained retrieval on four large benchmarks, Market-1501 \cite{Zheng2015Scalable}, CompCars \cite{yang2015large}, CUB-200-2011 \cite{wah2011caltech} and CARS196 \cite{cars196}.
\end{itemize}

\section{Related Work}
\paragraph{Person Re-identification.} Person re-identification is a challenging task, due to the appearance variations caused by lighting, pose, viewpoint, or occlusion changes. The previous works to tackle this task can be divided into two categories, namely, non-deep learning methods and deep learning methods. Non-deep learning methods extract hand-crafted features \cite{zhao2013person,liao2015person}, especially color and texture descriptors, then use distance metric learning \cite{koestinger2012large, liao2015person} to further capture discriminative factors. Deep learning methods \cite{yi2014deep,ahmed2015improved,varior2016siamese} learn image representations and distance metrics simultaneously in an end-to-end manner. However, the performance of most of the previous works were limited due to insufficient utilization of attribute and regional features.

\paragraph{Fine-grained Recognition.}  Deep learning methods have shown significant improvement on the fine-grained recognition. A bilinear model is used to learn the interacted feature of two independent CNNs, which performs superbly in bird classification \cite{lin2015bilinear}. Key parts of the object are utilized by some methods to extract subtle features \cite{zhang2014part, branson2014bird}. But those methods heavily rely on manually pre-defined parts, so some methods proposed to automatically discover critical parts in weak supervised way \cite{zhang2016picking, fu2017look}. A recent proposed attention model \cite{liu2017localizing} used attributes to help crop discriminative parts, but the hard-cropped parts may not be the optimal and the attribute features were not sufficiently utilized. OPAM \cite{peng2018object} proposes object-part attention model to localize object and discriminative parts. FDL \cite{shu2016image} uses tailored fine-grained dictionaries to help image classification. Cascaded Part-Based System \cite{biglari2018cascaded} learns a part-based model for each category and a cascading scheme for fine-grained classification.

\paragraph{Attributes for Image Recognition.} In image recognition, attribute is a type of important auxiliary information \cite{lampert2009learning, layne2012person} and have been investigated in many works. Yutian \emph{et al.} \cite{lin2017improving}, Tetsu \emph{et al.} \cite{matsukawa2016person} train CNNs using attribute loss, while Chi \emph{et al.} \cite{su2016deep} use attributes triplet loss to fine-tune CNNs. They simply use the last layer features without further exploring the importance of different features and the relationship between attributes and category. The person dataset Market-1501 \cite{Zheng2015Scalable}, bird dataset CUB-200-2011 \cite{wah2011caltech} and car dataset CompCars \cite{yang2015large} are annotated with attribute labels, which facilitate the future research on attributes for fine-grained recognition. In this paper, we explore attention mechanism to select better attribute features.

\paragraph{Metric Learning for Fine-grained Image Retrieval.} Metric learning projects images to a high dimensional embedding space, the main goal is to narrow the distances between similar images while pushing away dissimilar images as far as possible in embedding space. \cite{Cui2016Fine, mining} take image  triplets as inputs and learn a similarity metric by adding triplet comparison constraints. Unfortunately, because of the massive training data ($\Theta(N^3)$), they need a carefully designed hard example mining method to choose triplets for accelerating converge speed in training phase. \cite{npair} defines constraints on all images in a batch. Some other works \cite{bier, hdc} ensemble multiple independent models, the sub-embeddings generated by previous models are primarily learned over easy samples and the latter sub-embeddings are mainly tuned on hard samples, leading to the state-of-the-art performance on various benchmarks with the drawback that training multiple models are time-consuming and take on too much memory.

\paragraph{Attention Mechanism.} The attention mechanism has recently been used in computer vision and natural language processing, such as image caption \cite{xu2015show}, machine translation \cite{bahdanau2014neural} and visual question answering \cite{xu2016ask,yang2016stacked}. Here we focus on soft attention mechanism which computes a weighted combination of the features to focus on important parts when performing a particular task. Within an image, the importance of different parts is different when comparing whether two images are from the same category. As mentioned above, attention models have been used in fine-grained recognition to crop the key parts. In \cite{liu2017end}, attention mechanism was firstly used in person re-identification by integrating a recurrent attention with the siamese model. Liming \emph{et al.} \cite{Zhao2017Deeply} performed human body partition to learn more robust representations, which is a kind of hard attention model. In this paper, we propose a novel Attribute-Aware Attention Model to generate attentions for global category features fusion and local attribute features fusion simultaneously.

\begin{figure}[!htb]
\centering
\subfigure[]{
\label{fig:attention1}
\includegraphics[width=1.0\linewidth]{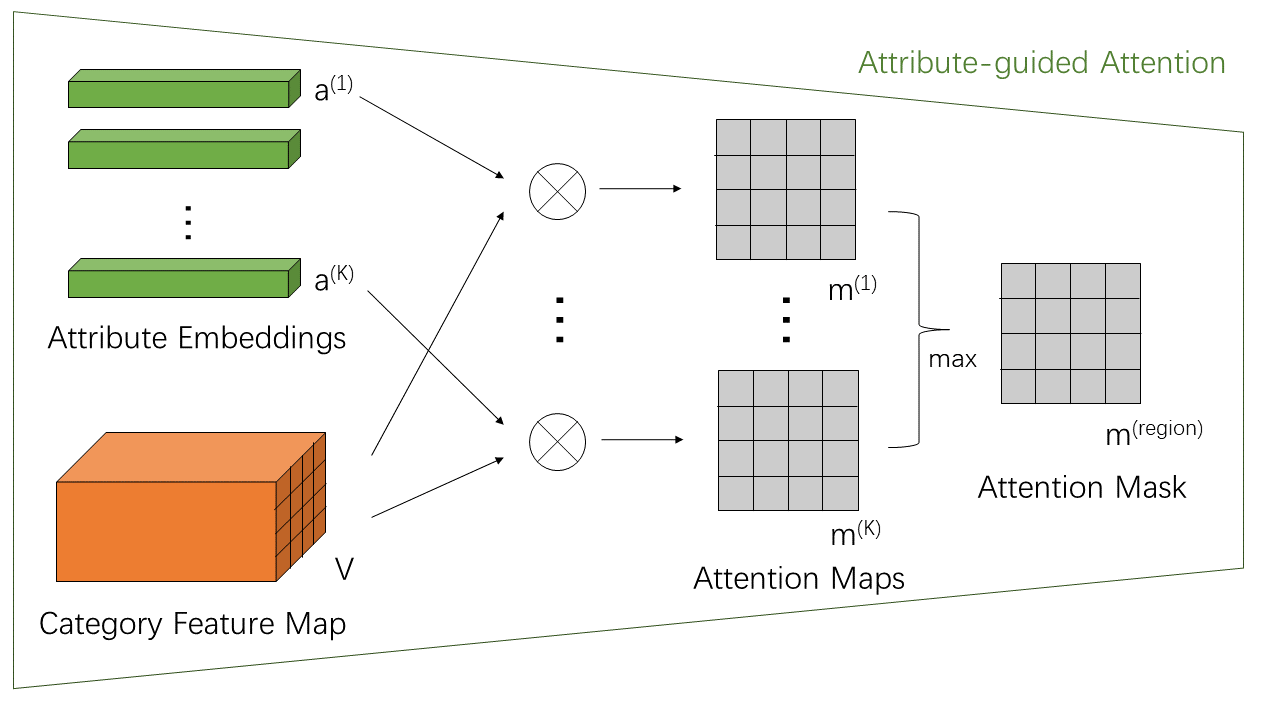}}
\subfigure[]{
\label{fig:attention2}
\includegraphics[width=1.0\linewidth]{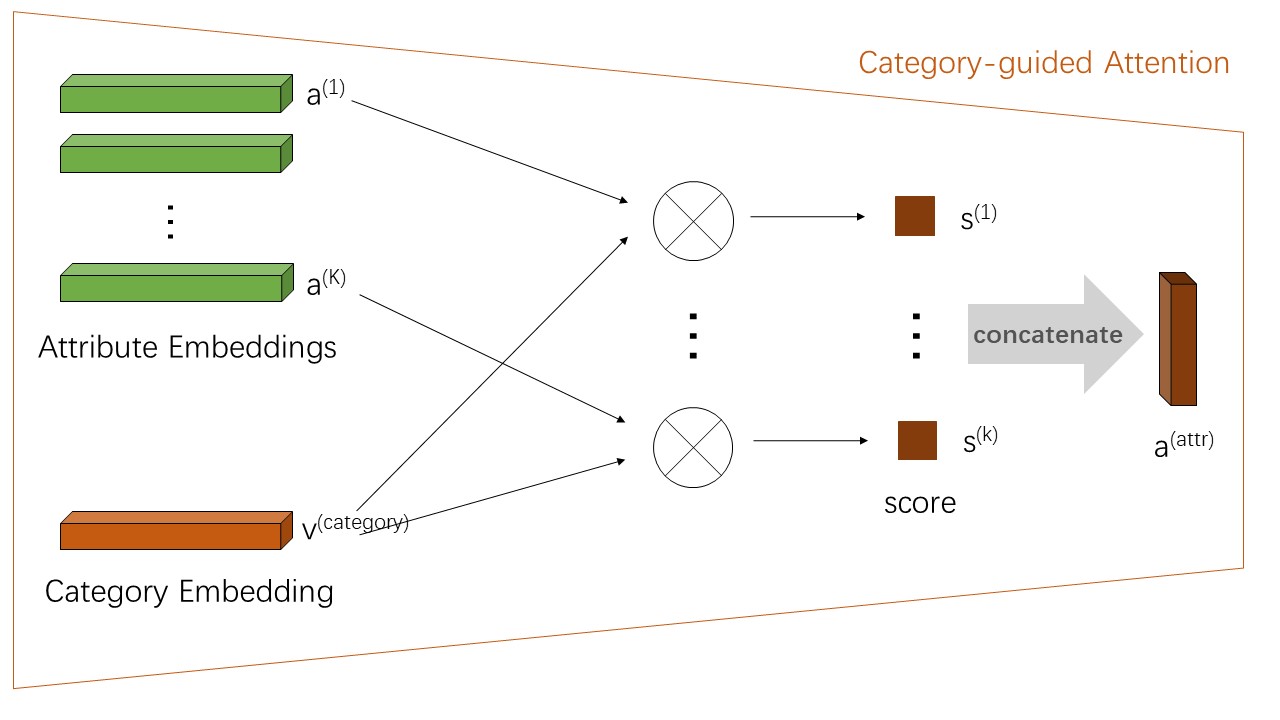}}
\caption{Attribute-Category Reciprocal Attention module, including two components: (a) attribute-guided attention module, (b) category-guided attention module.}
\label{fig:attention}
\end{figure}

\section{Attribute-Aware Attention Model}
In this paper, we propose an end-to-end Attribute-Aware Attention Model for fine-grained representation learning. We firstly introduce two branches in the network extracting attribute feature and category feature separately, then explain the attribute-guided attention module (the green one in figure \ref{fig:attention}) and category-guided attention model (the orange one in figure \ref{fig:attention}), which can help each other to select better attribute feature and category feature simultaneously, and finally present $A^3M$ architectures for person re-identification and image recognition tasks.

\subsection{Initial Category and Attribute Recognition}
\subsubsection{Shared CNN}
The shared CNN can be various networks, such as AlexNet \cite{Krizhevsky2012ImageNet}, VGG \cite{simonyan2014very}, ResNet \cite{he2016deep}, etc. And we take off the final fully connected layer. The shared CNN is pre-trained on ImageNet dataset \cite{russakovsky2015imagenet}. Given an input image, it is used to extract high-level visual features of local regions. Taking ResNet-50 for example, when the size of the input image is set as $224\times224\times3$, and the output feature map after the last pooling layer is in the dimension of $2048\times7\times7$. This feature map is shared by all the subsequent branches.

\subsubsection{Category Branch}
In the category recognition branch (the orange feature extraction branch in Figure \ref{fig:fig1}), after a convolution layer with $d$ $2048\times1\times1$ kernels, the shared feature map is transformed to category-related feature map in the dimension of $d\times h\times w$. Each $d$-dimensional vector in the feature map corresponds to a local region of the input image, for instance, the last feature map's size from ResNet-50 is $2048\times 7\times 7$, thus we obtain $L=49$ local feature vectors, simply represented as $V=[v_1,v_2,\cdots,v_{L}]$ where $v_l\in\mathbb{R}^{d}, l=1,2,\cdots,L$. After a global average pooling layer, we get the category embedding $v^{(category)}\in\mathbb{R}^{d}$. The prediction of category classification is given by the fully connected layer with softmax activation, and cross-entropy loss is used for training.
\begin{equation}
\hat{p}^{(category)}=softmax(W^{(c)}v^{(category)}+b^{(category)}),
\end{equation}
\begin{equation}\label{loss1}
\mathcal{L}^{(category)}=-\sum_{j=1}^{C^{(category)}}p_j^{(category)}\log{\hat{p}_j^{(category)}},
\end{equation}
$\hat{p}^{(category)}$ is the predicted probability, $W^{(c)}\in\mathbb{R}^{C^{(category)}\times d}$, $b^{(category)}\in\mathbb{R}^{C^{(category)}}$ are the weight matrix and bias vector of the fully connected layer, and $C^{(category)}$ is the number of categories. Let $t$ be the ground-truth category label, so that $p_{j\neq{t}}^{(category)}=0$ for all $j$ and $p_t^{(category)}=1$. The learned category embedding $v^{(category)}$ contains global information for image recognition.

\vspace{-0.03cm}
\subsubsection{Attribute Branch}
As shown in the green branch in Figure \ref{fig:fig1}, for the $k$-th attribute of the object ($1\leq k\leq K$, $K$ is the number of attributes), we feed the shared feature map into convolution layer with $d$ $2048\times1\times1$ kernels and a pooling layer to obtain attribute embedding vector $a^{(k)}\in\mathbb{R}^{d}$ for every attribute. Similar to category recognition, we use softmax loss for attribute classification.
\vspace{0.05cm}
\begin{equation}
\hat{p}^{(k)}=softmax(W^{(k)}a^{(k)}+b^{(k)}),
\end{equation}
\begin{equation}\label{loss2}
\mathcal{L}^{(k)}=-\sum_{j=1}^{C^{(k)}}p_j^{(k)}\log{\hat{p}_j^{(k)}},
\end{equation}
where $W^{(k)}\in\mathbb{R}^{C^{(k)}\times d}$, $b^{(k)}\in\mathbb{R}^{C^{(k)}}$ are the weight matrix and bias vector, and $C^{(k)}$ is the number of the $k$-th attribute categories. $\hat{p}^{(k)}$ is the predicted probability, and $p_j^{(k)}$ is the target probability. Let $t$ be the target attribute label, so that $p_t^{(k)}=1$ and $p_j^{(k)}=0$ for all $j\neq t$. During training, the loss function will force the $k$-th attribute embedding $a^{(k)}$ paying more attention to the regions related to the $k$-th attribute.  All $K$ attribute embeddings are learned by the way mentioned above.

\subsection{Attribute-Category Reciprocal Attention}\label{ACRA}
Based on the category feature map and attribute embeddings, we design an attribute-category reciprocal attention module to use them effectively for fine-grained representation learning. The intuition is to select regions in category feature map which are most relevant to intrinsic attributes and to select attributes most related to category. This is a reciprocal process with the help of each other. The proposed $A^3M$ consists of two components: attribute-guided attention module and category-guided attention module, as shown in Figure \ref{fig:attention}.

\paragraph{Attribute-guided Attention for Regional Features Selection.}
We give an overview of the attribute-guided attention module in Figure \ref{fig:attention1}. For each attribute, the attribute-guided attention module takes the category feature map $V$ and $K$ attribute embeddings ($a^{(1)}$, $\cdots$, $a^{(K)}$) as input, and produces an attention map for regional features. The intuition here is to select the local regions involved with intrinsic attributes rather than the background relevant parts. The $k$-th attribute-guided attention weights are given as follow:
\begin{equation}
m^{(k)}=\sigma(V^Ta^{(k)}),
\end{equation}
 where $\sigma(x)=1/(1+e^{-x})$ is sigmoid function. The generated attention weight mask $m^{(k)}\in\mathbb{R}^L$ reflects the correlations between the $L$ local regions and the $k$-th attribute. There are $K$ attributes, so we get $K$ attention maps. After merging them via max-pooling, we get $m^{(region)}=\max(m^{(1)},m^{(2)},\cdots,m^{(K)})$ as the final attribute-guided attention map . The values in the resulting attention map $m^{(region)}\in\mathbb{R}^L$ are high in selected regions and low in other regions. The regional features are multiplied by the attention weights and summed to produce the category representation $f^{(region)}\in\mathbb{R}^d$,
\begin{equation}
f^{(region)}=\frac{1}{L}Vm^{(region)}.
\end{equation}

\paragraph{Category-guided Attention for Attribute Features Selection.}
The category-guided attention module is illustrated in Figure \ref{fig:attention2}. With the $K$ attribute embeddings $A=[a^{(1)},\cdots,a^{(K)}]$ and the category embedding $v^{(category)}$, the category-guided attention weights are computed similarly to attribute-guided attention,
\begin{equation}
s^{(attr)}=\sigma(A^Tv^{(category)}).
\end{equation}
The $K$ attributes features are fused via category-guided attention weighting:
\begin{equation}
f^{(attr)}=\frac{1}{K}As^{(attr)}.
\end{equation}
The resulting attribute representation $f^{(attr)}\in\mathbb{R}^d$ selects attributes which are most relevant to category recognition.

\begin{figure}[!htp]
\begin{center}
   \includegraphics[width=0.9\linewidth]{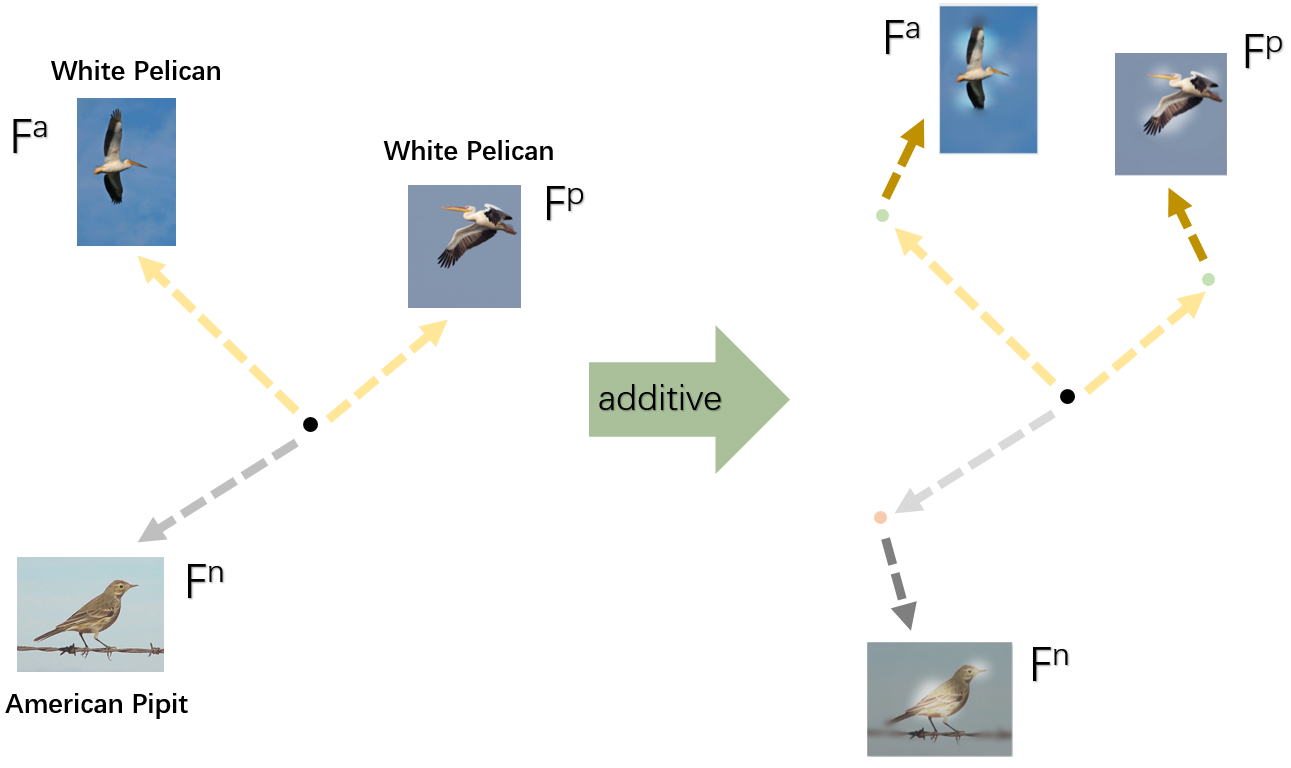}
\end{center}
   \caption{2D example for the addition of category feature and attribute feature. The global category feature embedding of the three images corresponds the left one, with the help of adding attribute feature, final feature can learn a better representation.}
\label{fig:additive}
\end{figure}

\subsection{Final Category Recognition}
The final feature is the addition of the weighted category feature and attribute feature, $f^{(final)}=f^{(region)}+f^{(attr)}$, and it contains information of both category and intrinsic attributes after reciprocal selection. As shown in figure \ref{fig:additive}, we can learn better feature embedding through attention models and local attribute features help to refine the global category feature, enabling the similar images to have smaller distances and dissimilar images have larger distances (right part of figure \ref{fig:additive}). Our goal in this paper is fine-grained visual recognition, thus the main category classification with cross-entropy loss is used for final representation learning,
\begin{equation}
\hat{p}^{(add)}=softmax(Wf^{(final)}+b),
\end{equation}
\begin{equation}\label{loss3}
\mathcal{L}^{(add)}=-\sum_{j=1}^{C^{(category)}}p_j^{(category)}\log{\hat{p}_j^{(add)}},
\end{equation}
$\hat{p}^{(add)}$ is the predicted category probability, $W\in\mathbb{R}^{C^{(category)}\times d}$, $b\in\mathbb{R}^{C^{(category)}}$ are the weight matrix and bias vector. By combining all the loss \eqref{loss1}, \eqref{loss2} and \eqref{loss3}, the final loss function for end-to-end training in our method is
\begin{equation}
\mathcal{L}^{(final)}=\mathcal{L}^{(additive)}+\alpha \mathcal{L}^{(category)}+\beta\frac{1}{K}\sum_{k=1}^{K}\mathcal{L}^{(k)},
\end{equation}
where $\alpha$ and $\beta$ are the trade-off parameters.


\begin{table*}[!htp]
\renewcommand\arraystretch{1.1}
\centering
\caption{Comparison with state-of-the-art on Market-1501 dataset. We also provided the results of baseline models. - means that no reported results are available.}
\label{table:market-soa}
\begin{tabular}{l|c|c|c|c|c}
\Xhline{1.0pt}
Methods    & rank-1   & rank-5   & rank-10  & rank-20 & mAP   \\ \hline
CAN \cite{liu2017end} & 60.3  & - & - & -   & 35.9 \\
Gated S-CNN \cite{varior2016gated}  & 65.88  & - & - & -   & 39.55\\
CRAFT-MFA+LOMO \cite{chen2017person} & 71.8  & - & - & -   & 45.5 \\
Re-ranking \cite{zhong2017re} & 77.11 & - & - & -   & 63.63 \\
ResNet-50 (I+V) \cite{zheng2016discriminatively}  & 79.51 & 90.91 & 94.09 & 96.23 & 59.87 \\
Body-Parts-Fusion \cite{Li2017Learning} & 80.31  & - & - & -   & 57.53 \\
Part-Aligned \cite{Zhao2017Deeply}  & 81.0 & 92.0 & 94.7 & 96.4 & 63.4 \\
SSM \cite{bai2017scalable}  & 82.21 & - & - & -   & 68.80 \\ 
JLML \cite{ijcai2017-305} & 83.9 & - & - & -   & 64.4 \\
Attr-Id \cite{lin2017improving} & 84.29 & 93.20 & 95.19 & 97.00 & 64.67 \\ \hline
Baseline-1   & 73.90 & 87.68 & 91.54 & 94.80 & 47.78 \\
Baseline-2   & 80.07 & 91.06 & 94.12 & 96.08 & 59.13 \\
$A^3M$ w/o attr-label   & 85.21 & 94.57 & 96.53 & 97.92 & 66.42 \\
$A^3M$  & \textbf{86.54} & \textbf{95.16} & \textbf{97.03} & \textbf{98.30} & \textbf{68.97} \\ \hline
\end{tabular}
\end{table*}

\section{Experiments}
We conducted experiments on a person re-identification dataset, two fine-grained classification datasets and two image retrieval datasets.

\subsection{Datasets}
\subsubsection{Person Re-Identification}
\paragraph{Market-1501}
Market-1501 \cite{Zheng2015Scalable} is a large person re-identification dataset, containing 32,668 bounding box images of 1,501 persons captured by 6 cameras. The dataset is split into 751 identities for training and 750 identities for testing. \cite{lin2017improving} annotated 27 attributes for each category, we follow \cite{lin2017improving} and use the attributes annotated Market-1501 dataset for experiments.

\subsubsection{Fine-grained Image Classification}
\paragraph{CUB-200-2011}
CUB-200-2011 \cite{wah2011caltech} is a well-known fine-grained benchmark dataset for birds classification. It contains 11,788 images of 200 classes. Each image is annotated with 15 part locations, 312 binary attributes, and 1 bounding box. Here we classify the attributes to 28 multi-class attributes for experiments. The object bounding box is used in our experiments.

\paragraph{CompCars}
CompCars \cite{yang2015large} is a large-scale car dataset for fine-grained categorization and verification. We use the categorization subset which contains 431 car models with a total of 36,456 training images and 15,627 test images. 
Each car model is annotated with five attributes, namely, maximum speed, displacement, number of doors, number of seats, and type of car.

\subsubsection{Fine-grained Image Retrieval}
\paragraph{CUB-200-2011}
For image retrieval, we don't use the attributes information mentioned above, the first 100 classes are for training (5,864 images) and the rest of classes are for testing (5,924 images).

\paragraph{CARS196}
CARS196 dataset \cite{cars196} has 196 classes of cars with 16,185 images, where the first 98 classes are for training (8,054 images) and the other 98 classes are for testing (8,131 images).

\subsection{Implementation Details}
We implemented our model using Keras \cite{chollet2015keras} deep learning framework. The embedding dimension was set as $d=512$ for person re-id and image classification experiments, for image retrieval, we set the last FC layers to be 128 and 384 for fair comparison with other methods. We empirically set the value of the trade-off parameters as $\alpha=0.5$, $\beta = 0.5$. As mentioned above, the shared CNN was pre-trained on ImageNet dataset \cite{russakovsky2015imagenet}, and then we fine-tuned the pre-trained model on the target datasets. During fine-tuning, SGD optimizer with momentum was used to update the weights. We trained the model for 60 epochs with batch size of 32. The initial learning rate was set as 0.001 and changed to 0.0001 in the last 10 epochs. We used a momentum of $\mu=0.9$ and weight decay of $5e^{-4}$. All the experiments were conducted on a NVIDIA Pascal TITAN X GPU.

\begin{figure}[!htp]
\begin{center}
   \includegraphics[width=1.0\linewidth]{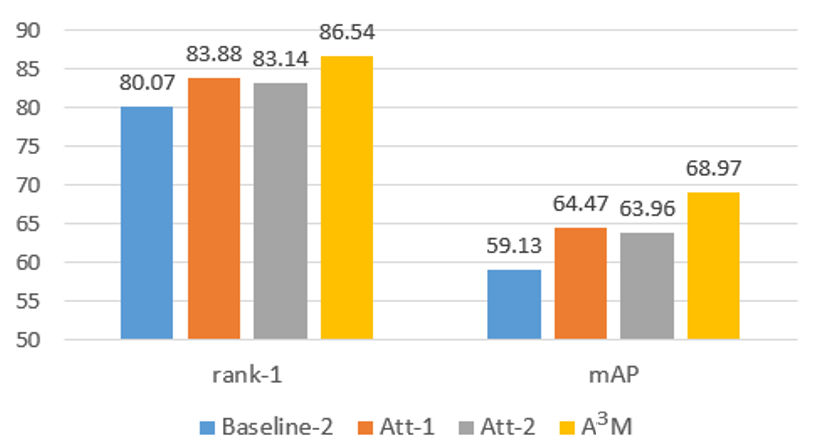}
\end{center}
   \caption{Systematic evaluation of the attention modules on Market-1501 dataset. The shared CNN architecture is ResNet-50.}
\label{fig:market-attention-compare}
\end{figure}

The size of model input in different dataset was a little different. In Market-1501 dataset, input images was resized to $448\times224$, the input size was $448\times448$ in CompCars dataset and $224\times224$ for other datasets. Random crop and random flip were used as data augmentation. 

We adopt the mean average precision (mAP) \cite{bai2017scalable} as evaluation metrics for person re-identification, and the normalized mutual information or NMI metric \cite{nmi} to measure clustering quality for image retrieval tasks. For image classification, accuracy is used for metric.

\begin{figure*}[htb]
\centering
\subfigure[]{
\label{fig:visualization1}
\includegraphics[width=0.9\linewidth]{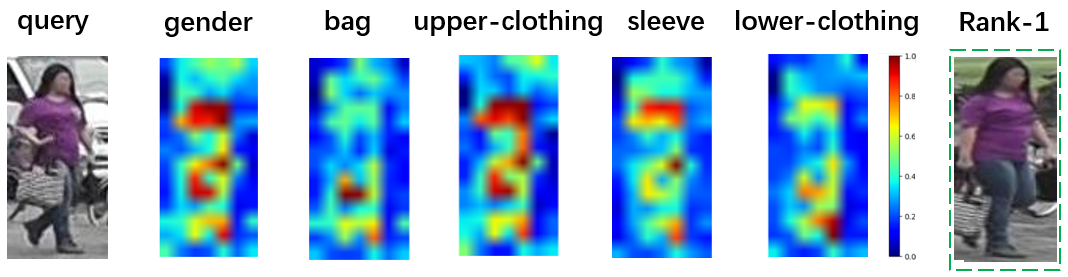}}
\subfigure[]{
\label{fig:visualization2}
\includegraphics[width=0.9\linewidth]{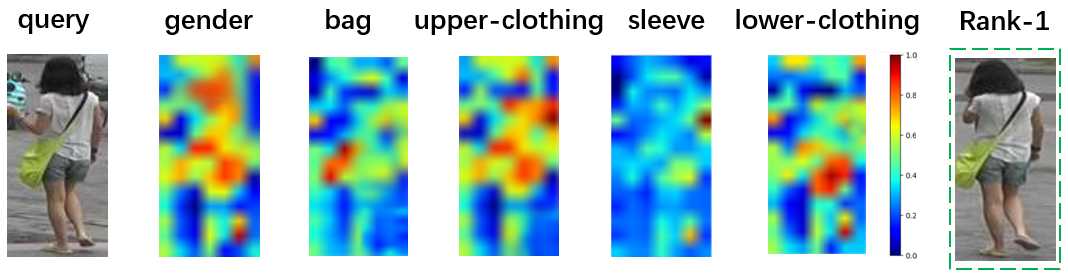}}
\caption{Visualization of some attention maps. The first column is the query image, and the rest columns show the attribute-guided attention maps (from left to right: gender, carrying bag, upper-clothing color, sleeve length, lower-clothing color), red means high value while blue means low. (a) and (b) are from different query persons. The last column shows the Rank-1 result of $A^3M$.}
\label{fig:visualization}
\end{figure*}

\begin{figure}[!htp]
\begin{center}
   \includegraphics[width=1.0\linewidth]{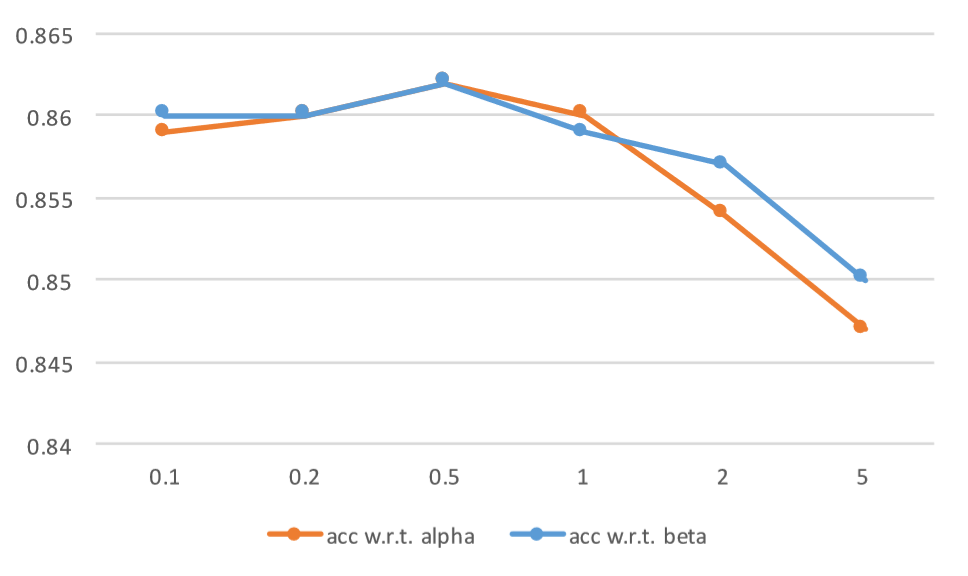}
\end{center}
   \caption{Classification accuracy w.r.t $\alpha$/$\beta$ on CUB-200-2011 dataset.}
\label{fig:hyper-param}
\end{figure}

\subsection{Evaluations on Person Re-identification}
\paragraph{Comparison with the state-of-the-art methods.}
We compare our method with other state-of-the-art methods, and the person re-identification results are listed in Table \ref{table:market-soa}. With ability to learn representation automatically, deep learning methods achieve impressive results. \cite{lin2017improving} has explored the usage of attributes by a multi-task CNN, which is much similar to our Baseline-2. We come up with two attention models which are identity-attribute reciprocal to further improve the usage of attribute features and
regional features. JLML \cite{ijcai2017-305} is designed to share identity label constraints by allocating each attribute branch with loss function separately, however, in real world applications, pedestrians may take off their coats at any time due to the weather, or one will help his or her partner to hold the handbag, separately train the local attribute feature to distinguish each person is not enough. The attention model in $A^3M$ and pair-wise addition in final representation can help to combine attributes and identity in a more flexible manner, making CNN more robust to pick crucial local attribute feature. $A^3M$ outperforms these state-of-the-art methods, achieves 2.25\% and 4.30\% improvements on rank-1 and mAP compared with \cite{lin2017improving}. On the one hand, the previous works are designed with treating all the features equally; on the other hand, $A^3M$ pays more attention to the important parts rather than the background and integrates attribute features with attention weights.

\paragraph{Ablation Study.}
We construct 2 baseline models to demonstrate the effectiveness of proposed $A^3M$ on Market-1501. Baseline-1 is a CNN pre-trained on ImageNet \cite{russakovsky2015imagenet} and fine-tuned to predict the person category on the target dataset. Baseline-2 is a simplification of proposed $A^3M$ by replacing the attention weights with equal weights. The results with ResNet-50 as the base network are shown in Table \ref{table:market-soa}. It can be seen that Baseline-1 obtains a decent result, and Baseline-2 achieves better performance with the auxiliary information of attributes. Our $A^3M$ exceeds both baselines by a significant margin. This illustrates that the attention mechanism can capture the important parts of spatial features and attribute features. For $A^3M$ w/o attr-label, we don't use attributes' label information in loss to supervise the CNN, that is to say, our $A^3M$ can learn local attribute feature in an unsupervised manner, which outperforms Baseline-2 by a large margin.

\begin{figure*}[!htp]
\vskip 0.4 cm
\begin{center}
   \includegraphics[width=1.0\linewidth]{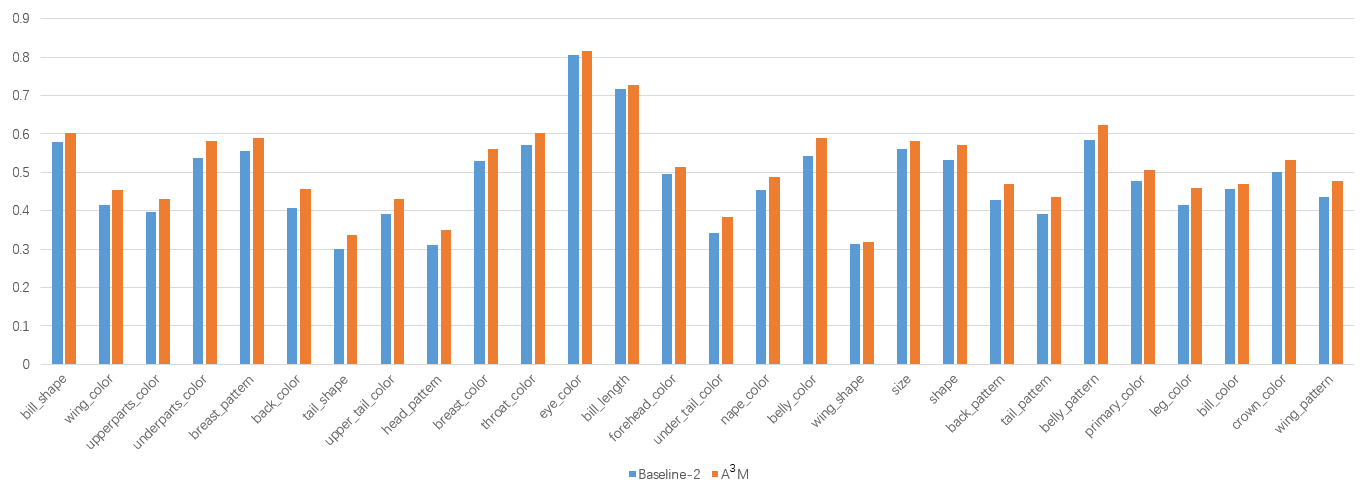}
\end{center}
   \caption{Attribute classification accuracy on CUB-200-2011 dataset. }
\label{fig:cub-attr}
\end{figure*}

In order to see whether both attention modules are effective, we conduct two more experiments to evaluate each attention module. Att-1 set the category-guided attention weights to equal values, which means that attribute features are fused via averaging. Att-2 uses equal values as attribute-guided attention weights, indicating the regional features are fused by average pooling. From Figure \ref {fig:market-attention-compare}, the model only uses attribute-guided attention or category-guided attention outperforms Baseline-2 which doesn't use any attention module. This shows that either attention module is beneficial to person re-identification. In addition, $A^3M$ defeats either Att-1 or Att-2 by a large margin, indicating that the usage of both attention modules is helpful to each other and obtained the best results.

\paragraph{Visualization of Attention Maps.}
We also visualize the attribute-guided attention maps for two example images in test set. Figure \ref{fig:visualization} shows two query images (the first column) with the selected 5 attribute-guided attention maps (the middle 5 columns) of regional features. It can be seen that in different attribute-guided attention maps, high values distribute in different regions. The corresponding relations are: gender - body region, bag - bag region, upper-clothing - upper body region, sleeve length - sleeve region and lower-clothing - lower body region. Through the attention model, $A^3M$ can select crucial local attribute cues such as bag and lower-clothing to help person re-identification. Nevertheless, attribute sleeve is not easy to learn well. The visualized results verify our idea that the attention mechanism can select regions most related to the intrinsic attributes for re-identification.

\subsection{Evaluations on Image Classification}

\begin{table}[!htp]
\renewcommand\arraystretch{1.0}
\centering
\caption{Experimental results on CUB-200-2011 dataset. w/ means using attributes or parts annotation during training.}
\label{table:cub}
\begin{tabular}{l|c|c|c}
\Xhline{1.0pt}
Methods                   & Attribute    & Parts    & Acc($\%$)             \\ \hline
MG-CNN \cite{wang2015multiple}    &    &     &  83.0            \\
DeepLAC \cite{zhang2016picking}    &    &     &  84.5            \\
Bilinear-CNN \cite{lin2015bilinear}               &    &     &  85.1             \\
RA-CNN \cite{fu2017look} &     &  & 85.3 \\
Part RCNN \cite{zhang2014part}   &       &  w/    & 76.4 \\
PS-CNN \cite{huang2016part}  &     & w/       & 76.6 \\
PN-CNN \cite{branson2014bird} & &  w/ & 85.4 \\
Image+Parts+Attribute \cite{liu2017localizing}      &   w/     &     & 85.5            \\ \hline
Baseline-1         &     &    & 82.3             \\
Baseline-2       & w/    &    & 84.3             \\
$A^3M$      & w/   &    &  \textbf{86.2}             \\ \hline
\end{tabular}
\end{table}

\begin{table}[!htp]
\renewcommand\arraystretch{1.0}
\centering
\caption{Experimental results on CompCars dataset. w/ means using attributes annotation during training.}
\label{table:car}
\begin{tabular}{l|c|c}
\Xhline{1.0pt}
Methods                   & Attribute        & Acc($\%$)             \\ \hline
AlexNet  \cite{yang2015large}               &        & 81.9             \\
Overfeat \cite{yang2015large}              &     & 87.9             \\ 
GoogLeNet \cite{yang2015large}              &     & 91.2           \\ \hline
Baseline-1               &         &  91.8             \\
Baseline-2               & w/        & 94.1             \\
$A^3M$              & w/       & \textbf{95.4}           \\ \hline
\end{tabular}
\end{table}

\paragraph{Comparison with State-of-the-Art Methods.}
We conduct experiments on two benchmark fine-grained classification dataset, namely, CUB-200-2011 \cite{wah2011caltech} and CompCars \cite{yang2015large}. The results of our model and previous state-of-the-art methods are summarized in Table \ref{table:cub} and Table \ref{table:car}. From the results shown, our model shows competitive performance with the best accuracy. This verifies the effectiveness of our model.

It can also be seen that the proposed method exceeds both baselines significantly, with a gain of at least 1.3\% in accuracy over baselines. Our $A^3M$ which integrates both attention modules further improves result, indicating that the attribute-guided attention module and category-guided attention module help to each other to obtain a better performance.

\begin{table*}[!htp]
\renewcommand\arraystretch{1.0}
\centering
\caption{Recall@K(\%) and NMI(\%) on CUB-200-2011 for image retrieval}
\label{table:cub-retrieval}
\begin{tabular}{l|c|c|c|c|c|c|c}
\Xhline{1.0pt}
Methods & Network & Dim & R@1 & R@2 & R@4  & R@8 & NMI   \\ \hline
Npairs \cite{npair} & GoogLeNet & 128  & 47.2  & 58.9  & 70.2 & 80.2 & 56.2  \\
HDC \cite{hdc} & GoogLeNet & 384 & 53.6 & 65.7  & 77.0 & 85.6 & -  \\
BIER \cite{bier}  & GoogLeNet & 512 & 55.3 & 67.2 & 76.9 & 85.1 & - \\ \hline
HDC & InceptionBN  & 384 & 60.0 & 71.8 & 81.5 & 88.6 & 65.6 \\ 

Baseline-1 & InceptionBN  & 384 & 59.1 &  70.7 & 81.5 & 88.8 & 65.6 \\
$A^3M$ & InceptionBN & 384 & \textbf{61.2} &  \textbf{72.4} & \textbf{81.8} & \textbf{89.2} & \textbf{66.4}  \\ \hline
\end{tabular}
\end{table*}

\begin{table*}[!htp]
\renewcommand\arraystretch{1.0}
\centering
\caption{Recall@K(\%) and NMI(\%) on CARS196}
\label{table:car-retrieval}
\begin{tabular}{l|c|c|c|c|c|c|c}
\Xhline{1.0pt}
Methods & Network & Dim & R@1 & R@2 & R@4  & R@8 & NMI   \\ \hline
Npairs \cite{npair} & GoogLeNet & 128  & 71.1  & 79.7  & 86.5 & 91.6 & 62.7  \\
HDC \cite{hdc} & GoogLeNet & 384 & 73.7 & 83.2  & 89.5 & 93.8 & -  \\
BIER \cite{bier}  & GoogLeNet & 512 & 78.0 & 85.8 & 91.1 & 95.1 & - \\ \hline
HDC & InceptionBN  & 384 & 79.2 & 87.3 & 92.2 & \textbf{95.5} & 66.8 \\ 
Baseline-1 & InceptionBN  & 384 & 78.5 & 86.8 & \textbf{92.3} & 95.4 & 66.8 \\ 
$A^3M$ & InceptionBN & 384 & \textbf{80.0} &  \textbf{87.5} & \textbf{92.3} & \textbf{95.5} & \textbf{67.7}  \\ \hline
\end{tabular}
\end{table*}

\paragraph{Analysis of hyper-parameters.}
To further evaluate the impact of the trade-off parameters $\alpha$ and $\beta$, we show in Figure \ref{fig:hyper-param} the result variation on CUB-200-2011 dataset. It can be seen that the performance is gradually improved when $\alpha$ or $\beta$ increase from 0.1 to 0.5, however, further increasing $\alpha$ makes the performance decrease. The performance is not sensible to the hyper-parameters, so we can easily choose a proper trade-off parameter to obtain a good performance in practice.

\paragraph{Evaluation of Attribute Classification.}
We test attribute classification on the testing set of CUB-200-2011 dataset, and the results are listed in Figure \ref{fig:cub-attr}. The proposed $A^3M$ and Baseline-2 are compared. The overall accuracy of $A^3M$ is higher than that of Baseline-2, with an significant increase, indicating the attribute-category reciprocal attention module not only benefits to category classification, but also helps to learn a more discriminative attribute classification model.

\subsection{Evaluations on Image Retrieval}
To ensure the fairness of the comparison, we reproduce the HDC method based on the InceptionBN. HDC \cite{hdc} trains an ensemble of diverse models of different hardness levels, achieving Recall@1 at 60.0\% and 79.2\% on CUB-200-2011 and CARS196, Table \ref{table:cub-retrieval} quantifies the advantages of $A^3M$ on CUB-200-2011. $A^3M$ avhieves 61.2\% for Recall@1, which is significantly better than all the previous state-of-the-art methods including the reproduced HDC. Table \ref{table:car-retrieval} reports the results on CARS196, where $A^3M$ achieves improvements around 1\% over the baseline-1 measured by Recall@\{1,2\}.

\section{Conclusion}
In this paper, we propose a novel Attribute-Aware Attention Model to learn fine-grained representation for the tasks of person re-identification, fine-grained classification and retrieval. In our model, both global category features and local attribute features are learned, and an attribute-category reciprocal attention module is used to select the most important category regional features and attribute features with the help of each other. Experimental results have shown the effectiveness of our method in fine-grained image recognition problems.

\section*{Acknowledgement}
This work was supported in part by the National Natural Science Foundation of China under Grant 61671027 and the National Key Basic Research Program of China  under Grant 2015CB352303.
